\documentclass[10pt,twocolumn,letterpaper]{article}

\usepackage{iccv}
\usepackage{times}
\usepackage{epsfig}
\usepackage{graphicx}
\usepackage{amsmath}
\usepackage{amssymb}
\usepackage{graphics}
\usepackage{adjustbox}
\usepackage{xcolor}
\usepackage{footmisc}

\usepackage[breaklinks=true,bookmarks=false]{hyperref}

\iccvfinalcopy 

\def\iccvPaperID{5} 

\ificcvfinal\fi

\begin{document}
	
	\title{LIP: Learning Instance Propagation for Video Object Segmentation}
	\author{Ye Lyu\\
		University of Twente\\
		{\small y.lyu@utwente.nl}
		\and
		George Vosselman\\
		University of Twente\\
		{\small george.vosselman@utwente.nl}
		\and
		Gui-Song Xia\\
		Wuhan University\\
		{\small guisong.xia@whu.edu.cn}
		\and
		Michael Ying Yang\\
		University of Twente\\
		{\small michael.yang@utwente.nl}
	}

	\maketitle
	\ificcvfinal\thispagestyle{empty}\fi
	\begin{abstract}
		In recent years, the task of segmenting foreground objects from background in a video, \ie video object segmentation (VOS), has received considerable attention.
		In this paper, we propose a single end-to-end trainable deep neural network, convolutional gated recurrent Mask-RCNN, for tackling the semi-supervised VOS task. 
		We take advantage of both the instance segmentation network (Mask-RCNN) and the visual memory module (Conv-GRU) to tackle the VOS task. The instance segmentation network predicts masks for instances, while the visual memory module learns to selectively propagate information for multiple instances simultaneously, which handles the appearance change, the variation of scale and pose and the occlusions between objects. After offline and online training under purely instance segmentation losses, our approach is able to achieve satisfactory results without any post-processing or synthetic video data augmentation. Experimental results on DAVIS 2016 dataset and DAVIS 2017 dataset have demonstrated the effectiveness of our method for video object segmentation task.
	\end{abstract}
	
	\begin{figure*}[h]
		\begin{center}
			\includegraphics[width=1.0\linewidth]{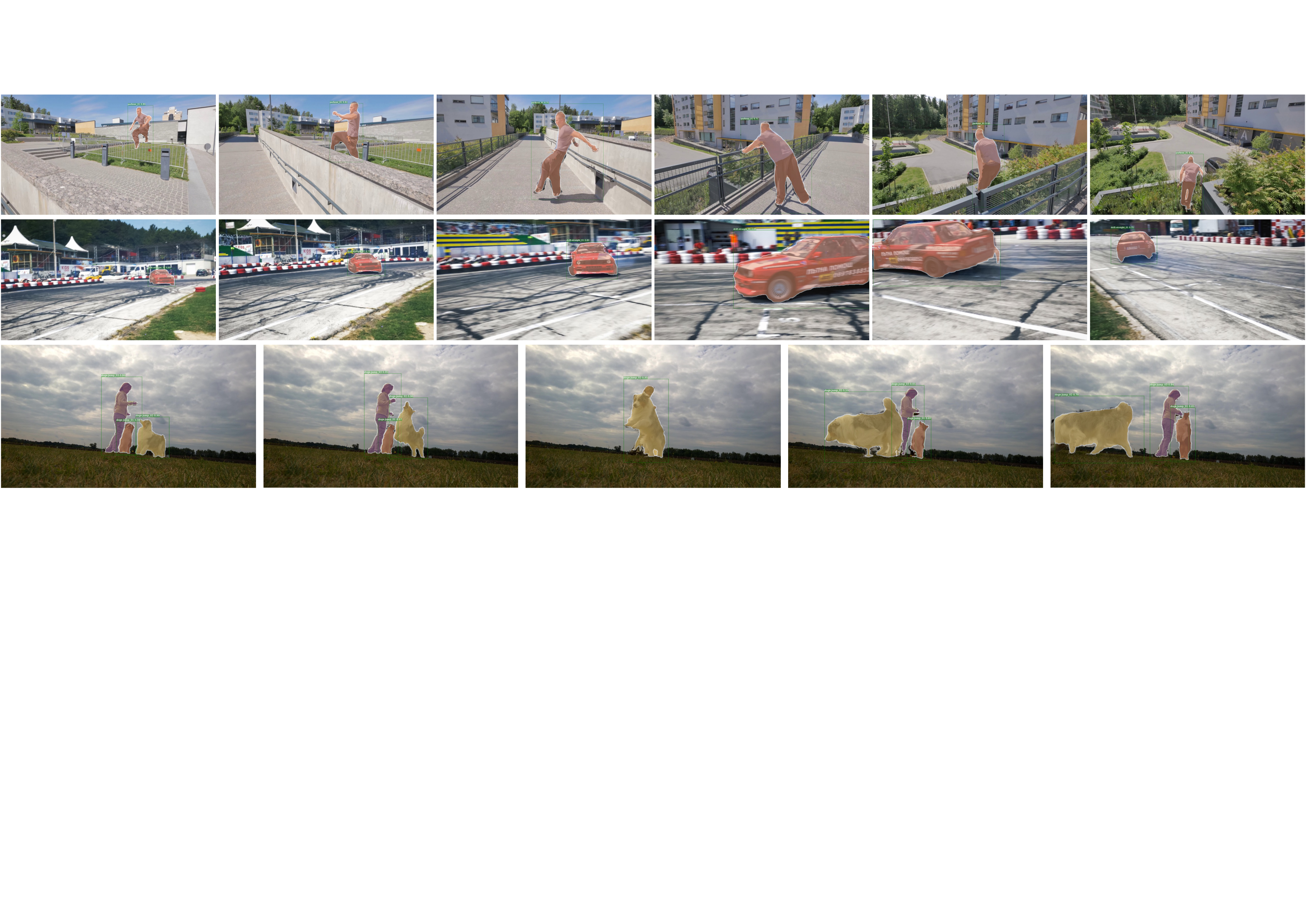}
		\end{center}
		\caption{Example predictions by our method from DAVIS~\cite{davis_2017, davis16} dataset. Top row: Parkour sequence. An example of large appearance change over time. One every 20 frames shown of 100 in total. Middle row: Drift-straight sequence. An example of large scale and pose variation over time. One every 10 frames shown of 50 in total. Bottom row: Dogs-jump sequence. An example of occlusions between objects. One every 5 frames shown of first 20.}
		\label{fig:eg_vos}
	\end{figure*}
	
	\section{Introduction}
	Video object segmentation (VOS) aims at segmenting foreground objects from background in a video with coherent object identities. Such visual object tracking task serves for many applications including video analysis and editing, robotics and autonomous cars. Compared to the video object tracking task in bounding box level, this task is more challenging as pixel level segmentation is more detailed description of an object. 
	
	The VOS task is defined as a semi-supervised problem if ground truth annotations are given for the first several frames. It is otherwise an unsupervised problem if no annotation is provided. The ground truth annotations are masks that mark the objects that need to be tracked through the whole video. In our work, we focus on semi-supervised video object segmentation task, where the ground truth annotations are provided only for the first frame. 
	
	There are several challenges that make VOS a difficult task. First, both the appearance of the target objects and the background surroundings may change significantly over time. Second, there could be a large pose and scale variation over time. Third, there could be occlusions between different objects, which hinder the performance of tracking. Examples of the above three challenges are shown in Fig.~\ref{fig:eg_vos}.
	A notable and challenging dataset for the VOS task is the DAVIS 2016 dataset \cite{davis16}, which is designed for single-object video segmentation. Later the DAVIS 2017\cite{davis_2017} is brought out focusing on segmentation of multiple video objects. Both of the datasets are provided with mask annotations of extremely high accuracy.
	
\begin{figure*}[t]
		\begin{center}
			\includegraphics[width=1.0\linewidth]{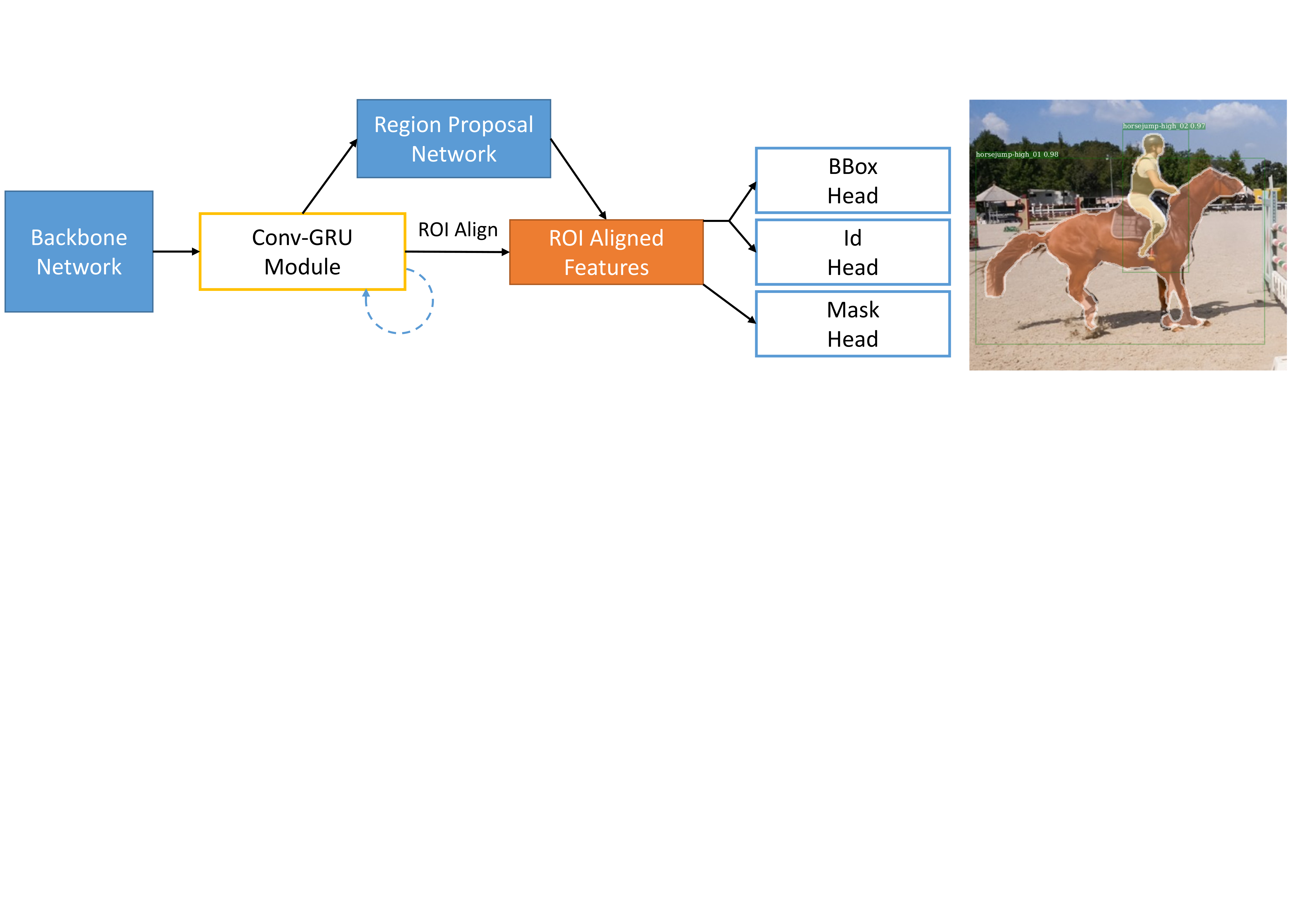}
		\end{center}
		\caption{Overall model structure. The backbone network distills useful features from each input image. The features are then sent to Conv-GRU module (visual memory) for feature propagation. The output features from Conv-GRU module are utilized by region proposal network for proposal generation. Multiple heads finally take the ROI aligned features for video object segmentation. An example output is shown on the right, including bounding boxes, id predictions and object segments. The class of an instance is named by video sequence name plus object index.}
		\label{fig:model_structure}
	\end{figure*}

	Most of the current methods for the VOS task, such as VPN~\cite{VPN}, MSK~\cite{VOSfromStatic} and RGMP~\cite{fast_mask_prop_VOS},  are based on the pixel level mask propagation. However, those methods fail to give a coherent label within an instance. 
	In this paper, we introduce a single end-to-end trainable network to predict masks on instance level, namely the convolutional gated recurrent Mask-RCNN. It integrates instance segmentation network (Mask-RCNN~\cite{MaskRCNN}) with visual memory module (Conv-GRU~\cite{convgru}). Instance segmentation network is designed for foreground object segmentation, which is extended with visual memory for foreground object segmentation in a video. The incorporated visual memory helps to propagate information across frames to handle the appearance change, the pose and scale variation and the occlusions between objects. Our network gives a coherent label to a detected instance and assigns one label to only one detected instance. The model structure is shown in Fig. \ref{fig:model_structure}.
	
	Our \textbf{Contributions} are: 
	\begin{itemize}
		\item We propose a novel convolutional gated recurrent Mask-RCNN to learn instance propagation (\textbf{LIP}) for video object segmentation (VOS) task. 
		Our model simultaneously segments all the target objects in the images.
		\item We design a single end-to-end trainable network for VOS task, enabling both mask propagation in the long term and bottom-up path augmentation.
		\item A strategy to successfully train the model for VOS task has been brought out. All the training processes are guided by the instance segmentation losses only. 
	\end{itemize}

	\section{Related work}
	In this section, we will discuss some relevant work.
	
	\noindent
	\textbf{Object detectors}. Object detection starts with box level prediction and has a great improvement over the years. Single-stage detectors~\cite{YOLO,YOLO9000,ssD,DssD,focalloss} have faster running speed while two-stage networks~\cite{fastrcnn,fasterrcnn} are more accurate in general. Later, Mask-RCNN~\cite{MaskRCNN} merges object detection with semantic segmentation by combining Faster-RCNN~\cite{fasterrcnn} and FCN~\cite{FCN}, which form a conceptually simple, flexible yet effective network for instance segmentation task. Mask-RCNN network is suitable for instance segmentation on static images, but lacks the ability for temporal inference. Our work is to further extend Mask-RCNN with Conv-GRU module to solve video object segmentation task.
	
	\noindent
	\textbf{Recurrent neural networks (RNNs)}. RNNs~\cite{RNN_hopfield,RNN_rumelhart} are widely used for tasks with sequential data, such as image captioning~\cite{RNN_visual_description}, image generation~\cite{RNN_ImageGeneration} and speech recognition~\cite{RNN_speech}. The key for RNNs is the hidden state, which selectively accumulates information from current input and the previous hidden state over time. However, RNN has its limitation as it fails to propagate information for a long sequence due to the problem of gradient vanishing or explosion in training~\cite{RNN_vanishing, RNN_difficulty}. Two RNN variants, LSTM~\cite{LSTM} and GRU~\cite{GRU} are more effective for the long term prediction by taking advantage of gating mechanism. To further encode spatial information, they are extended to Conv-LSTM~\cite{convlstm} and Conv-GRU~\cite{convgru} respectively and have been used for video prediction~\cite{convlstm_video_pred} and action recognition~\cite{convgru}.
	
	\noindent
	\textbf{Methods for VOS}. Conv-GRU has already been used for video object segmentation. It serves as visual memory in~\cite{VOSVisualMemory} and has been proved to boost the performance for VOS task. However, their model performs binary semantic segmentation only, which is not suitable for video object segmentation task with multiple objects. 
	
	VPN~\cite{VPN}, MSK~\cite{VOSfromStatic} and RGMP~\cite{fast_mask_prop_VOS} learn to propagate mask for the VOS task. VPN utilizes learnable bilateral filters to achieve video-adaptive information propagation across frames. MSK learns to utilize both current frame and mask estimation from the previous frame for mask prediction. RGMP utilizes the first frame and mask as reference for instant information propagation besides the usage of current frame and previous mask estimation. Both MSK and RGMP achieve good results, but they can only propagate information for instances one by one.
	
	Specially, OSVOS~\cite{OSVOS}, OSVOS+~\cite{OSVOS+} and OnAVOS~\cite{OnAVOS} tackle video object segmentation from static images, achieving temporal consistency as a by-product. They learn a general object segmentation model from image segmentation datasets and transfer the knowledge for video object segmentation. They all rely on additional post processing for better segmentation result. OnAVOS further applies online adaptation to continuously fine-tune the model, which is very time consuming.
	
	\cite{LucidDataDreaming_CVPR17_workshops} explores the benefits from in-domain training data synthesis with the labelled frames of the test sequences. ~\cite{fast_mask_prop_VOS} synthesizes video training data from static image dataset to add to limited video training samples. ~\cite{VidMatch_VOS,PLM_VOS} explore fast prediction without online training through matching based method. CINM~\cite{CNN_MRF_VOS} achieves good prediction by spatial-temporal post-processing based on results from OSVOS~\cite{OSVOS}. To handle the problem of long term occlusion, ~\cite{ReID_VOS, Joint_VOS} apply re-identification network to retrieve the missing objects, which complements their mask propagation methods. Recently, there are still many researches focusing on single-object video segmentation~\cite{MoNet, motion_guide_vos, local_sensitive_vos}, which are not easily transformed for video segmentation of multiple objects. MaskRNN~\cite{maskrnn} is another method for instance level segmentation, but it only predicts for one instance at a time. 
	The best results are achieved by ensemble of multiple specialized networks. PReMVOS~\cite{premvos} takes the 1st place of recent DAVIS2018 semi-supervised VOS task by utilizing complex pipeline with multiple specialized networks trained on multiple datasets.
	
	\section{Method}
	In this section, we first introduce the structure of our convolutional gated recurrent Mask-RCNN, which extracts and propagates information for multiple objects in a video. It is comprised of mainly three parts. They are the feature extraction backbone, the visual memory module and the prediction heads. The backbone network extracts features that are forwarded to visual memory module. The visual memory module then selectively remembers the new input features and forgets the old hidden states. On top of Conv-GRU, region proposal network (RPN), bounding box regression head, id classification head and mask segmentation head are constructed to solve the VOS task. The whole network is end-to-end trainable under the guidance of instance segmentation losses.
	
	\subsection{Mask-RCNN}
	Mask-RCNN~\cite{MaskRCNN} is one of the most popular framework designed for instance segmentation task. It is used for instance-wise object detection, classification and mask segmentation, which makes it naturally suitable for multiple video objects segmentation. Roles of different components in Mask-RCNN directly shift to fit VOS task as illustrated below.
	
\begin{figure*}[t]
		\begin{center}
			\includegraphics[width=1.0\linewidth]{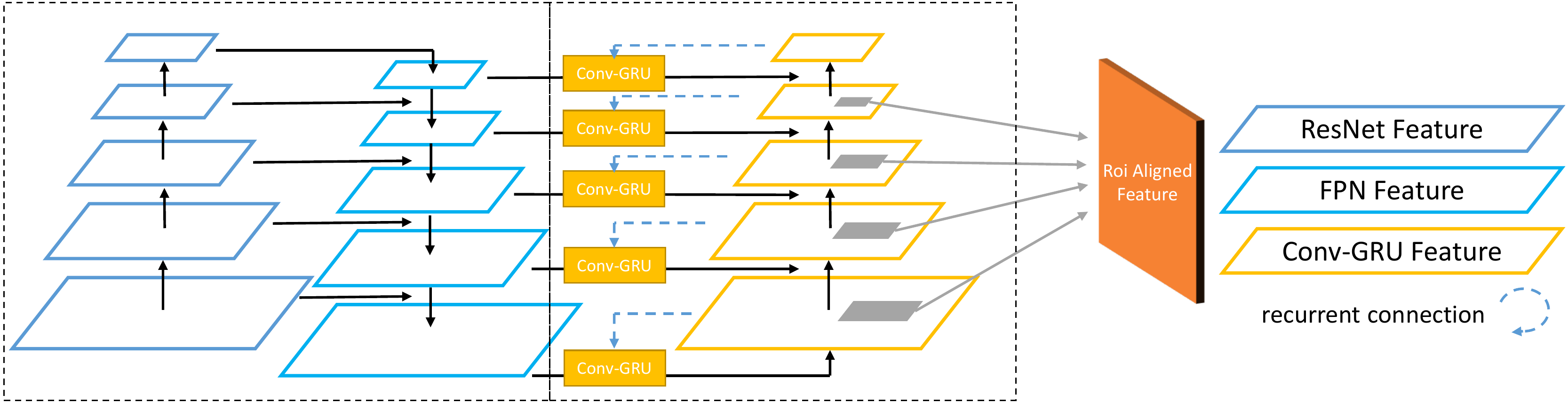}
		\end{center}
		\caption{Model structure details. The left black dashed box shows the ResNet101-FPN backbone structure. The right black dashed box shows the Conv-GRU module. Our network brings bottom-up path augmentation for output features in Conv-GRU module. The augmented output features are used for both RPN and the prediction heads. All 5 layers are utilized for multi-level RPN, but only 4 bottom layers are used for multi-level ROIs.}
		\label{fig:model_structure_details}
	\end{figure*}

	\textbf{Backbone:} The backbone network still serves to extract features from images, but more focused on generating useful features adaptively for gates of Conv-GRU module. ResNet101-FPN~\cite{ResNet,FPN} with group normalization~\cite{groupNorm} is used as our backbone network. Detailed structure is shown in Fig.\ref{fig:model_structure_details}.
	
	\textbf{RPN:} Mask-RCNN is known as a two stage instance segmentation network. Bounding boxes of general objects are proposed in the first stage, while classes and masks are predicted instance-wisely in the second stage. Such two stage framework adopts the same philosophy as the training stages of OSVOS~\cite{OSVOS}. For OSVOS, the network first learns to segment binary mask for general objects in a class-agnostic manner. Then it learns to segment specific objects during online training. In Mask-RCNN, RPN learns to reject background objects and to propose foreground objects in the first stage, which is also class-agnostic. It is in the second stage that classes and masks of different objects are determined.
	
	\textbf{Bounding box regression head:} This branch is used to refine the bounding box proposals. Each predicted box contains one object. The boxes serve to separate different objects in an image.
	
	\textbf{Classification head:} This branch is used to assign the object a correct class label. However, class type is unknown for VOS task. Instead, different objects are associated with different ids, which need to be predicted coherently in a video sequence. Classification branch is naturally transformed into an id classification branch.
	
	\textbf{Mask segmentation head:} This branch is used to extract a mask for each foreground object in the image, which is the main target of VOS task.
	
	Clearly, for the components in Mask-RCNN, there is a direct responsibility mapping from instance segmentation task to VOS task.
	
	\subsection{Convolutional gated recurrent unit}
	One difficulty for video object segmentation is the problem of long term dependency. The ground truth is provided only for the first frame, but the objects still need to be predicted after tens or hundreds of frames based on the ground truth from the first frame. The appearance of different objects in the videos may vary greatly and the objects sometimes get partially or even completely occluded, which makes coherent prediction more difficult.
	
	In order to handle the above problem, we utilize the convolutional gated recurrent unit, serving as a visual memory to handle appearance morphing and occlusion. The memory module learns to selectively propagate the memorized features and to merge them with the newly observed ones. The key role for Conv-GRU module is to maintain a good feature over time for prediction of region proposal, bounding box regression, id classification and mask segmentation. 
	
	Compared to the instance segmentation task, where each training batch is comprised of multiple randomly sampled images, the batch in temporal training has less variation as consecutive images from one sequence are highly correlated. This is similar to the problem of small batch size. To relieve such effect, we further replace the bias term in Conv-GRU with the group normalization (GN) layer, which are proved to give consistent performance across different batch sizes~\cite{groupNorm}:
	\begin{subequations} \label{eq_gru_gn}
		\begin{align}
		z_t & = \sigma(GN(W_{hz}*h_{t-1}+W_{xz}*x_t))\\
		r_t & = \sigma(GN(W_{hr}*h_{t-1}+W_{xr}*x_t))\\
		\hat{h_{t}} & = \Phi(GN(W_h*(r_t \odot h_{t-1})+W_x*x_t))\\
		h_t & = (1-z_t) \odot h_{t-1} + z_t \odot \hat{h_t}, 
		\end{align}
	\end{subequations}
	where $x_t$ is the input feature of time t, $h_t$ is the hidden state of time $t$. $z_t, r_t$ are update gate and reset gate respectively. 
	$W$ are convolutional filter parameters. $\sigma$ and $\Phi$ are sigmoid function and tanh function respectively.$*$ and $\odot$ denote the convolution operation and element-wise multiplication respectively.
	
	For each level of the feature pyramid network~\cite{FPN} (FPN), we create a corresponding Conv-GRU layer. The layers at different levels learn different transition functions for the hidden states. As bottom up path augmentation has been proved to be useful for instance segmentation~\cite{Liu_2018_CVPR}, we easily achieve it by down-sampling and addition operation with output features from multi-level Conv-GRU module. The structure is shown in Fig.~\ref{fig:model_structure_details}. The output features after path augmentation are used for RPN and prediction heads. Conv-GRU module is deliberately directly inserted after backbone network. In this way, information for both region proposal and instance prediction can be propagated through time.
	
	\subsection{Online inference}
	As our model predicts mask for each unique instance, there naturally exist constraints for prediction.
	
	\textbf{One maximum constraint}. For each instance, there should be at most one object detected. This constraint is achieved by selecting highest id prediction score.
	
	\textbf{Location continuity constraint}. If an instance is detected in the previous frame with high enough id prediction score, the location of the current detection should not be far from its previous location. To achieve this constraint, we suppress the prediction for the instance, whose boxes iou between consecutive frames is low.
	
	As probability for id prediction decays over time, we further apply a very light weighted fine-tuning process for the last linear layer of the id head during online prediction. If there exists a target object detected with a high enough id prediction score, its predicted bounding box is set as ground truth for fine-tuning the id head only. By saving and reusing intermediate tensors, the speed for fine-tuning is fast.
	
	\section{Training the network}
	In this section, we will describe our training strategy in detail.
	The training modality for video object segmentation can be divided into offline training and online training~\cite{LucidDataDreaming_CVPR17_workshops,VOSfromStatic}. During offline training, the model is trained with the training set only. During online training, the model is fine-tuned with the first frame from the test set. As the class types of the test set are not known and objects may never be seen during offline training, online fine-tuning is necessary to help the model to generalize better for test set. 
	
	Our network needs both offline training and online training. During offline training stage, our network learns the features to differentiate all the object instances and learns to predict class-agnostic boxes and masks. During online training stage, our network is fine-tuned to differentiate objects for each test sequence and trained with boxes and masks in a class-specific manner.
	
	\subsection{Class-agnostic offline training}
	To provide our model with as much generality as possible, we apply class-agnostic training for bounding box and mask through the whole offline training process. Offline training for our model can be divided into two steps. First, our model is trained with instance segmentation dataset. This step is to provide our model with general object detection ability. Then, we train the model with video dataset to learn to propagate information over time for video object segmentation.
	
	\subsubsection{Pre-train on instance segmentation dataset}
	Pre-training on additional dataset is a common practice~\cite{OnAVOS,OSVOS,OSVOS+,Joint_VOS}. We initialize our model by pre-training on Microsoft COCO dataset~\cite{COCO}. Ms-COCO dataset has been widely used for object detection task. It targets common objects in context with annotations including boxes, classes and masks. By first training on Ms-COCO dataset, our model learns to extract useful features for general object detection. As the training is on static images, we set hidden states to be zeros without update for Conv-GRU module.
	
	After this step, our model gains general region proposal ability, general bounding box prediction ability and general object segmentation ability. Our model also learns to differentiate general objects by classes defined on Ms-COCO dataset.
	
	\begin{figure}[t]
		\begin{center}
			\includegraphics[width=0.745\linewidth]{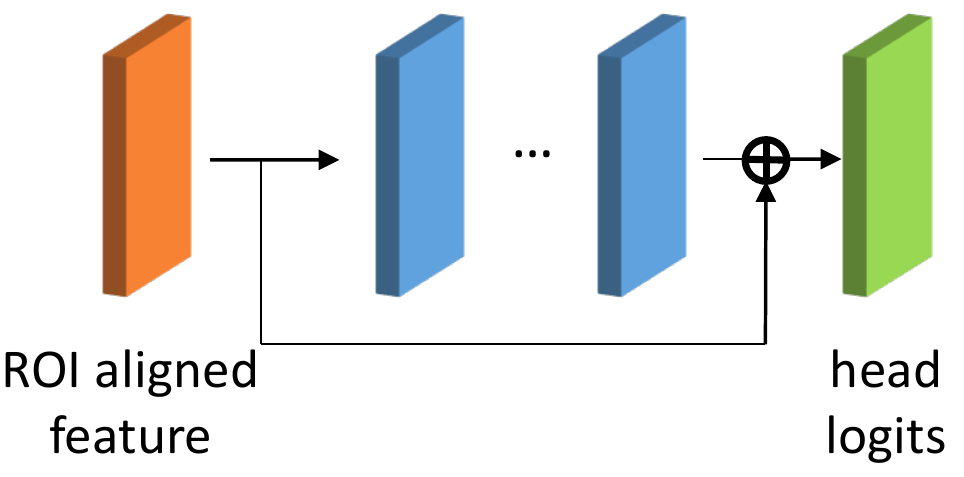}
		\end{center}
		\caption{Shortcut in prediction head. In order to let output from Conv-GRU module have more direct influence towards final prediction, we add a shortcut connection between ROI aligned feature and head logits by simple addition operation.}
		\label{fig:skip_con}
	\end{figure}
	
	\subsubsection{Fine-tune on VOS dataset} 
	\label{sec_fine_tune_vos}
	In this stage, we train all the modules except the backbone network. By fine-tuning our model on video object segmentation dataset, the Conv-GRU module learns to tune its gates to best propagate information. It should be noted that the class number has changed as the video object segmentation dataset does not share the class definition with instance segmentation dataset. Instead, we replace the last linear layer right before softmax layer in the class prediction head with a new one, which now predicts the ids in the dataset. The class prediction head turns into an id prediction head.
	
	The network is trained purely with instance segmentation losses. The different losses guide our model to have different abilities. The mask loss helps our model to propagate mask segmentation. The losses from id head and bbox head help our model to propagate information differently for each instance. Although the mask head and bbox head are trained in a class-agnostic manner, the id head and bbox head provide a chance to learn to propagate class-specific information.
	
	To facilitate the information propagation, we further add a shortcut connection between the ROI aligned feature and the head logits as shown in Fig.~\ref{fig:skip_con}.
	\begin{figure}[t]
		\begin{center}
			\includegraphics[width=0.9\linewidth]{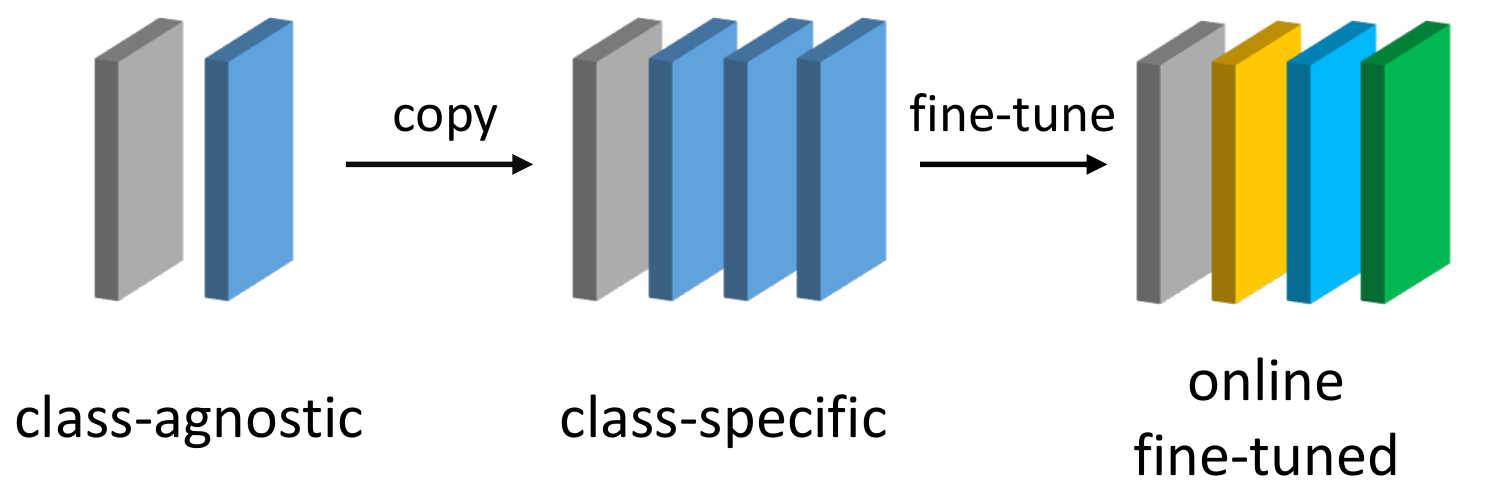}
		\end{center}
		\caption{Transforming class-agnostic weights to class-specific weights. During online fine-tuning, the class-agnostic bounding box and mask predictions are altered to class-specific. The rectangles are weights in the last linear layer of bbox head or the last convolutional layer of mask head. The grey color marks weights for background and the blue for foreground. Foreground weights are copied for each foreground instance to be fine-tuned uniquely.}
		\label{fig:agno_to_spec}
	\end{figure}

	\begin{figure*}[htbp]
		\begin{center}
			\includegraphics[width=0.95\linewidth]{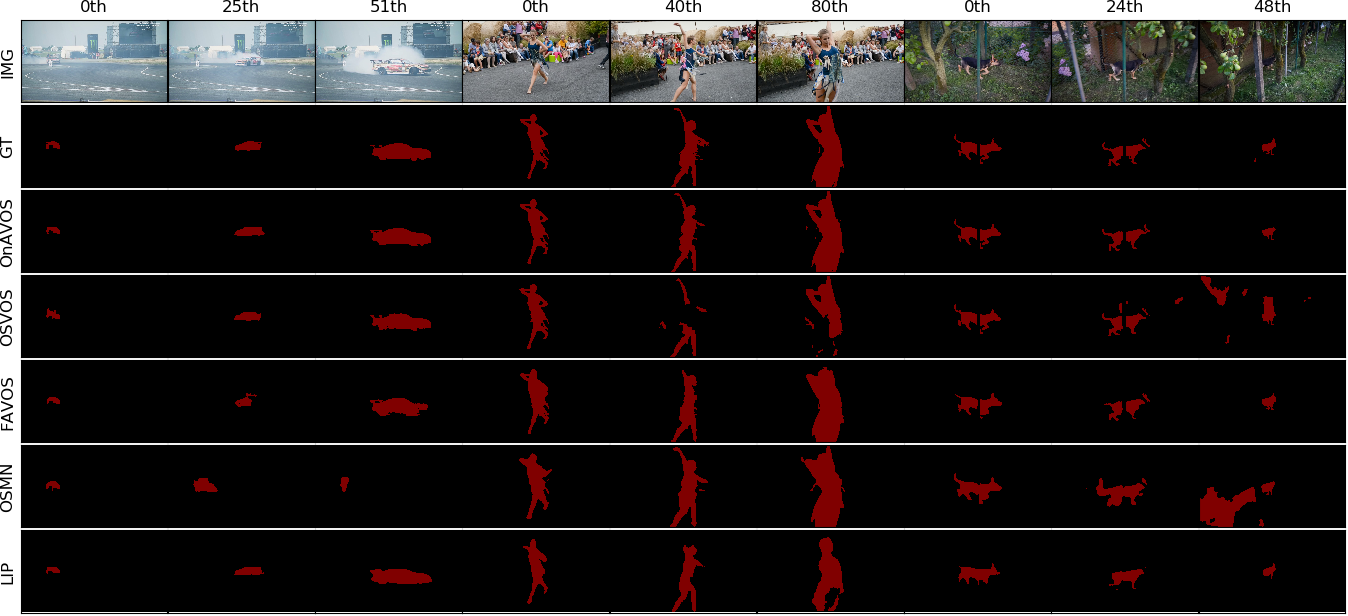}
		\end{center}
		\caption{Qualitative results comparison of OnAVOS~\cite{OnAVOS}, OSVOS~\cite{OSVOS}, FAVOS~\cite{fast_track_vos}, OSMN~\cite{efficient_vos} and LIP on DAVIS 2016 dataset~\cite{davis16}. The index of each image in a sequence is shown on the top.}
		\label{fig:davis16}
	\end{figure*}
	
	\subsection{Class-specific online fine-tuning}
	As the instances in test sequences are not the same as in training sequences, the last linear layer in id head needs to be re-initialized and trained to differentiate instances in the current sequence. We replace the last linear layer in the same way as in section~\ref{sec_fine_tune_vos}. We also adopt focal loss~\cite{focalloss} for id head to balance the training for multiple instances.
	
	During online fine-tuning, the parameters in backbone network and Conv-GRU module are frozen to keep the learned propagation property. All other parts are fine-tuned for the new objects in test image. The class-agnostic prediction in mask head and bbox head are altered to be class-specific in order to have less competition for different instances. The process is illustrated in Fig.~\ref{fig:agno_to_spec}.
	
	\section{Experiments}
	To test how our model learns to propagate instance information in a long term sequence, we evaluate our model on both DAVIS 2016~\cite{davis16} and DAVIS 2017~\cite{davis_2017} datasets, which contain video sequences of high quality and accurate mask annotations of objects. DAVIS 2016 dataset focuses on single-object video segmentation. It has 30 training and 20 validation videos. As an extension to DAVIS 2016 dataset, DAVIS 2017 dataset brings 30 more video sequences for training set and 10 more for validation set. It also provides another 30 sequences for testing. As DAVIS 2017 dataset focuses on multiple object segmentation, it has been re-annotated for each individual target object.
	
	\subsection{Implementation Details}
	Our model is implemented with PyTorch~\cite{pytorch} library. A Nvidia Titan X (Pascal) GPU with 12GB memory is used for experiments. Details of convolutional gated recurrent Mask-RCNN are shown below. 
	
	\noindent
	\textbf{Model structure.} Our backbone network is a ResNet101-FPN~\cite{ResNet,FPN} with group normalization~\cite{groupNorm}. ResNet101 is initialized with weights pre-trained on Imagenet~\cite{imagenet}. In Conv-GRU module, the channel number of each hidden state is 256. Kernels of all convolutions in Conv-GRU are of size $3\times3$ with 256 filters. 
	We apply multi-level RPN and multi-level ROIs for the network~\footnote{See supplementary material for more details.\label{ft:supp_ref}}. The ROI aligned feature resolution is $28\times28$ for mask head, and $7\times7$ for bbox head and id head. In all cases, we adopt image centric training~\cite{fastrcnn}.
	
	\begin{table*}[htp]
		\begin{center}
			\begin{tabular}{r|c|c|c|c|c|c|c|c|c|c}
				\hline
				Method             & OnAVOS  & FAVOS & OSVOS & LIP(Ours) & MSK & PML & SFL & OSMN & CTN & VPN\\
				\hline
				J\&F Mean$\uparrow$& 85.5    & 81.0  & 80.2  & 78.5      & 77.6& 77.4& 76.1& 73.5 & 71.4& 67.9\\
				\hline
				J Mean$\uparrow$   & 86.1    & 82.4  & 79.8  & 78.0      & 79.7& 75.5& 76.1& 74.0 & 73.5& 70.2\\
				J Recall$\uparrow$ & 96.1    & 96.5  & 93.6  & 88.6      & 93.1& 89.6& 90.6& 87.6 & 87.4& 82.3\\
				J Decay$\downarrow$& 5.2	 & 4.5	 & 14.9  & 0.05      & 8.9 & 8.5 & 12.1& 9.0  & 15.6& 12.4\\
				\hline
				F Mean$\uparrow$   & 84.9    & 79.5  & 80.6  & 79.0      & 75.4& 79.3& 76.0& 72.9 & 69.3& 65.5\\
				F Recall$\uparrow$ & 89.7    & 89.4  & 92.6  & 86.8      & 87.1& 93.4& 85.5& 84.0 & 79.6& 69.0\\
				F Decay$\downarrow$& 5.8     & 5.5	 & 15.0  & 0.06      & 9.0 & 7.8 & 10.4& 10.6 & 12.9& 14.4\\
				\hline
			\end{tabular}
		\end{center}
		\caption{Results on DAVIS 2016~\cite{davis16}. Left column shows different metrics. Up-arrow$\uparrow$ means the higher the better. Down-arrow$\downarrow$ means the lower the better. Methods are in descent order according to J\&F mean from left to right.}
		\label{tab:davis16}
	\end{table*}
	
	\begin{figure*}[htbp]
		\begin{center}
			\includegraphics[width=0.95\linewidth]{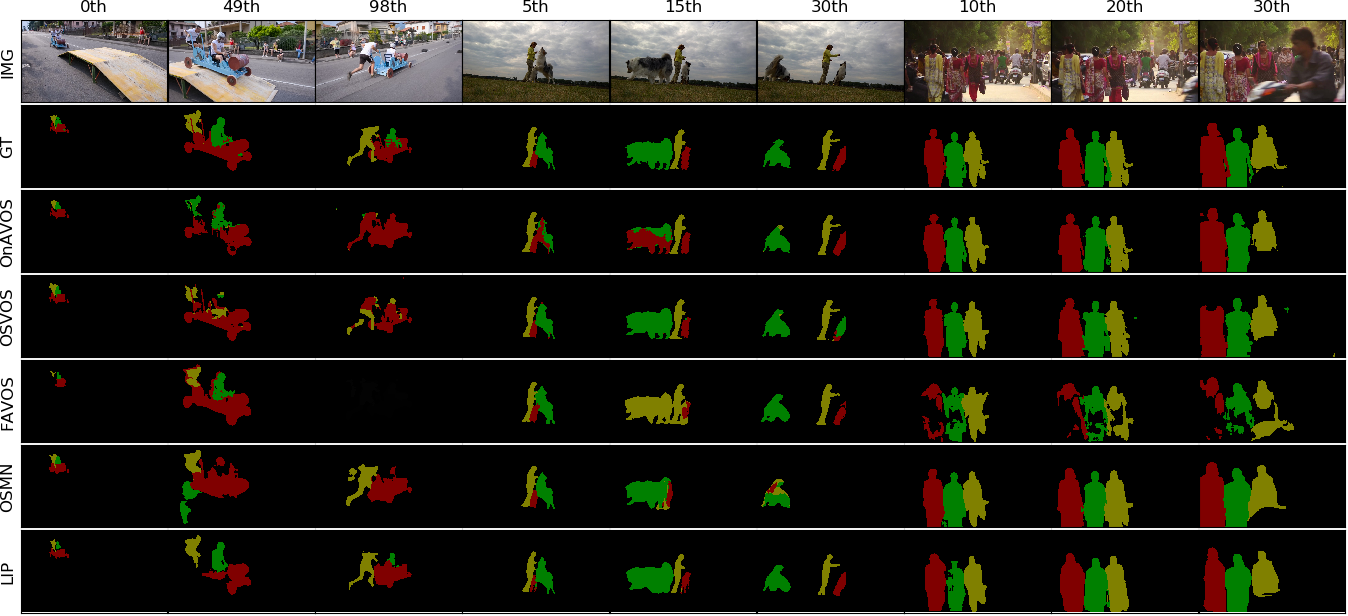}
		\end{center}
		\caption{Qualitative results comparison of OnAVOS~\cite{OnAVOS}, OSVOS~\cite{OSVOS}, FAVOS~\cite{fast_track_vos}, OSMN~\cite{efficient_vos} and LIP on DAVIS 2017 dataset~\cite{davis_2017}. The index of each image in a sequence is shown on the top.}
		\label{fig:davis17}
	\end{figure*}
	
	\noindent
	\textbf{Pre-train on Ms-COCO dataset.} For each image, we randomly scale it to have its shorter side equal to 1 of 11 different lengths: 640, 608, 576, 544, 512, 480, 448, 416, 384, 352, 320 and its longer size to be maximumly 1333. We sample 512 ROIs with foreground-to-background ratio 1:3. RPN adopts 5 aspect ratios (0.2, 0.5, 1, 2, 5) and 5 scales ($32^2$, $64^2$, $128^2$, $256^2$, $512^2$). The model is trained with stochastic gradient descent (SGD) for 270K iterations. We fix input hidden states to be zeros for Conv-GRU module, weight decay 0.0001, momentum 0.9. The initial learning rate is 0.02 and dropped by a factor of 10 at 210K and 250K. In the following cases, the configuration is kept the same unless otherwise stated.
	
	\noindent
	\textbf{Fine-tune on DAVIS dataset.}
	We generate ground truth (GT) bounding boxes from GT masks of DAVIS dataset. The width and height of the boxes are expanded by $10\%$ to prevent incomplete mask prediction caused by inaccurate box prediction. The sequences are randomly shuffled and scaled as in pre-training stage. As there is no causal reasoning in the task, we reverse each sequence for more training data. The backbone network is not trained to prevent over-fitting for DAVIS dataset. 128 ROIs are sampled from each image. The model is trained for 12K iterations with an initial learning rate of 0.002 and dropped by a factor of 10 at 8K and 10K. Due to the GPU memory limitation, it only allows to train with maximum recurrence of 4. We extend the length to 8 by stopping gradient back propagation between 4th and 5th frames.
	
	\noindent
	\textbf{Online fine-tuning.}
	The network is fine-tuned with the GT of the first image for maximally 1000 iterations with early stopping. If the loss for a prediction head is smaller than an empirically chosen threshold, the loss is ignored. If all the losses are ignored, we stop the training~\footref{ft:supp_ref}. We also stop the loss back-propagation in id head at its last fully connected layer, so the features to distinguish ids will not be affected by the newly initialized head. Focal loss~\cite{focalloss} is used to balance id training~\footref{ft:supp_ref}.
	
	\noindent
	\textbf{Online inference.}
	For each id, we select 10 detected objects that have id score above 0.2 and apply one maximum constraint to select the best candidate. For the location continuity constraint, we suppress the object instance that has IOU lower than 0.3 with the detection from previous frame if the previous id score is higher than 0.4. To relieve the id score from decaying over time, we apply fine-tuning for id head for maximally 500 iterations with early stopping~\footref{ft:supp_ref}.
	
	\begin{table*}[htp]
		\begin{center}
			\begin{tabular}{r|c|c|c|c|c}
				\hline
				Method              & OnAVOS & LIP(Ours)& OSVOS & FAVOS & OSMN\\
				\hline
				J\&F Mean$\uparrow$ & 65.4	 & 61.1     & 60.3  & 58.2  & 54.8\\
				\hline
				J Mean$\uparrow$    & 61.6   & 59.0     & 56.6  & 54.6  & 52.5\\
				J Recall$\uparrow$  & 67.4   & 69.0     & 63.8  & 61.1  & 60.9\\
				J Decay$\downarrow$ & 27.9	 & 16.0     & 26.1  & 14.1  & 21.5\\
				\hline
				F Mean$\uparrow$    & 69.1   & 63.2     & 63.9  & 61.8  & 57.1\\
				F Recall$\uparrow$  & 75.4   & 72.6     & 73.8  & 72.3  & 66.1\\
				F Decay$\downarrow$ & 26.6   & 20.1     & 27.0  & 18.0  & 24.3\\
				\hline
			\end{tabular}
		\end{center}
		\caption{Results on DAVIS 2017~\cite{davis_2017}. Left column shows different metrics. Up-arrow$\uparrow$ means the higher the better. Down-arrow$\downarrow$ means the lower the better. Methods are in descent order according to J\&F mean from left to right.}
		\label{tab:davis17}
	\end{table*}
	
	\subsection{Compare with other methods}
	We compare our method with some state-of-the-art methods on both the DAVIS 2016 benchmark 
	and the DAVIS 2017 benchmark \footnote{\url{https://davischallenge.org/}} by using standard evaluation metrics J and F~\cite{davis16,davis_2017}. The evaluation on DAVIS 2016 benchmark shows the performance for single-object video segmentation, while the evaluation on DAVIS 2017 benchmark shows the performance for video segmentation of multiple objects. It should be noted that our method does not apply any post-processing, but has been pre-trained on Ms-COCO dataset~\cite{COCO}. Among the several top methods, we remove CINM~\cite{CNN_MRF_VOS} and RGMP~\cite{fast_mask_prop_VOS} to avoid unfair comparison. CINM~\cite{CNN_MRF_VOS} is built upon OSVOS~\cite{OSVOS} and further adopts a refinement CNN and MRF for post-processing. The better initial prediction, the better its result. RGMP~\cite{fast_mask_prop_VOS} cannot be successfully trained with static image dataset and DAVIS dataset alone for mask propagation. It has created a large number of synthetic video training data from Pascal VOC~\cite{pascal_voc,pascal_voc1}, ECSSD~\cite{ecssd} and MSRA10K~\cite{msra10k} datasets. It is not fair to compare with RGMP as the quality of video training data are not the same and cannot be controlled. For DAVIS 2017 benchmark, we exclude PReMVOS~\cite{premvos} and OSVOS+~\cite{OSVOS+} as they both use multiple specialized networks in multiple processes to refine their results.
	
	For DAVIS 2016, we compare with OnAVOS~\cite{OnAVOS}, FAVOS~\cite{fast_track_vos}, OSVOS~\cite{OSVOS}, MSK~\cite{VOSfromStatic}, PML~\cite{blaz_fast_vos}, SFL~\cite{segflow_vos}, OSMN~\cite{efficient_vos}, CTN~\cite{ctn_vos} and VPN~\cite{VPN}. We detect multiple objects and evaluate in the way for single-object.
	Our method ranks the 4th among the compared methods as shown in Table~\ref{tab:davis16}. It should be noted that our results are better than FAVOS and OSVOS if they are without post-processing. FAVOS achieves J mean and F mean of 77.9\% and 76\% respectively without tracker and CRF~\cite{fast_track_vos}. OSVOS achieves J mean and F mean of 77.4\% and 78.1\% respectively without boundary snapping post-processing~\cite{OSVOS}. OnAVOS achieves J mean of 82.8\% without CRF post-processing~\cite{OnAVOS}.
	In addition, we compare our method with another visual memory (Conv-GRU) based VOS method~\cite{VOSVisualMemory}. Both of the methods are trained with additional image dataset, but we achieve 4.5\% gain in J\&F mean without optical flow and CRF post-processing.
	
	For DAVIS 2017, we compare LIP with OnAVOS~\cite{OnAVOS}, OSVOS~\cite{OSVOS}, FAVOS~\cite{fast_track_vos} and OSMN~\cite{efficient_vos} as shown in Table~\ref{tab:davis17}. LIP has relatively better performance as it is better at separating different instances and keeping coherent label within an instance. 
	
	Qualitative results on DAVIS 2016 and DAVIS 2017 are shown in Fig.~\ref{fig:davis16} and Fig.~\ref{fig:davis17}, respectively \footnote{More quantitative results and qualitative examples on DAVIS 2016 and DAVIS 2017 are shown in the supplementary material.}. Fig.~\ref{fig:davis16} shows that our LIP can track single object well on instance level and preserve good mask extent for an instance. OSMN~\cite{efficient_vos} and OSVOS~\cite{OSVOS} fail to keep the mask within an instance. 

	In Fig.~\ref{fig:davis17}, it is obvious that the information of an instance in our LIP helps segment multiple objects. 
	All the other methods either assign one label to multiple objects or assign multiple labels to one object, while LIP handles those issues better.
	
	\subsection{Ablation study}
\begin{figure}[htbp]
		\begin{center}
			\includegraphics[width=1.0\linewidth]{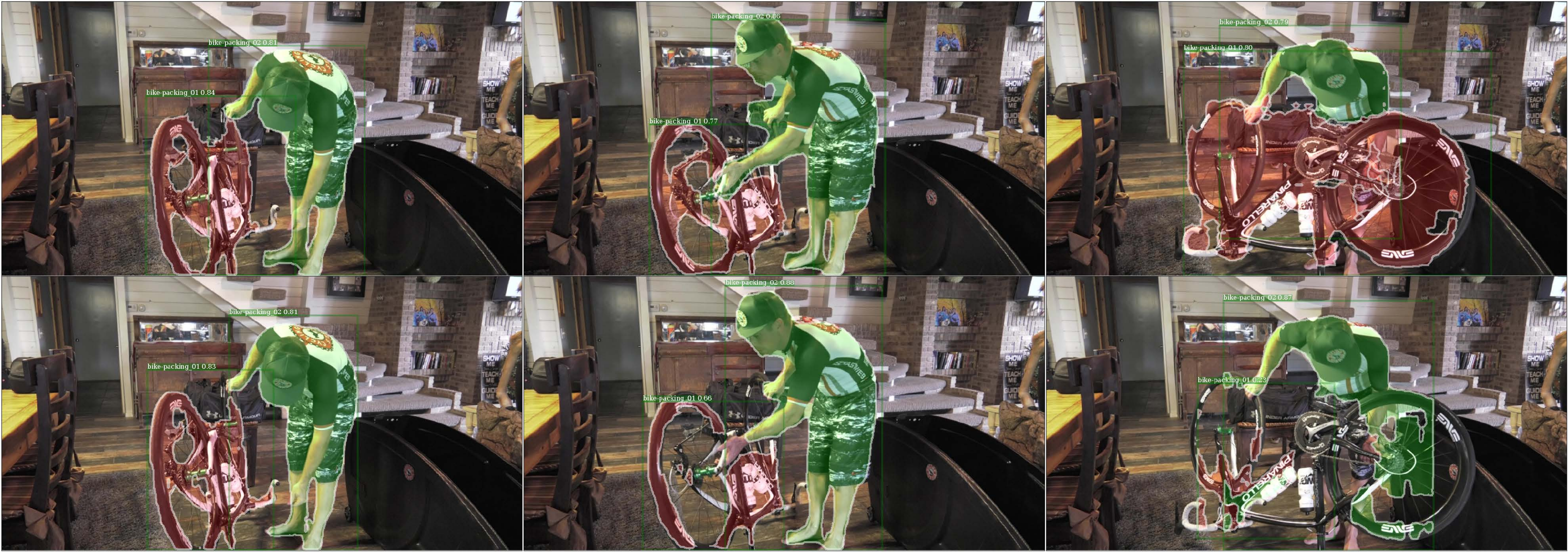}
		\end{center}
		\caption{A qualitative example of prediction with (top row) and without (bottom row) dynamic visual memory.}
		\label{fig:cmp}
	\end{figure}

	We perform ablation study on DAVIS 2017 dataset by comparing the model with and without dynamic visual memory as shown in Table~\ref{tab:ablation}. We first evaluate the static model by fixing input hidden states ($h_{t-1}$) to zeros for Conv-GRU module. This is Mask-RCNN with static Conv-GRU module and bottom up path augmentation. Fine-tuning on video dataset is done by training with static images only. The J\&F mean score is 59.2\%, which is 1 percent lower than the performance of OSVOS~\cite{OSVOS} with post-processing. The full version of our model is trained with dynamic video images. It reaches the best J\&F mean score of 61.1\%. The dynamic visual memory contributes as it learns to propagate masks. The static model lacks such property to handle large appearance change, as shown in Fig.~\ref{fig:cmp}.
	
	\begin{table}[htp]
		\begin{adjustbox}{width=\columnwidth,center}
			\begin{tabular}{c|c|c|c|c}		
				\hline
				Mask-RCNN &Conv-GRU  &J Mean&F Mean&J\&F Mean\\
				\hline
				\checkmark&input zero&56.9  &61.5  &59.2    \\
				&hidden states&&&\\
				\hline
				\checkmark&\checkmark&59.0  &63.2  &61.1    \\
				\hline
			\end{tabular}
		\end{adjustbox}
		\caption{Ablation study results on DAVIS 2017 dataset.}
		\label{tab:ablation}
	\end{table}
	
	\section{Conclusions}
	We have presented a single end-to-end trainable neural network for video segmentation of multiple objects. We extend the powerful instance segmentation network with visual memory for inference ability across time. Such design serves as an instance segmentation based baseline for VOS task. The newly designed convolutional gated recurrent Mask-RCNN learns to extract and propagate information for multiple instances simultaneously and achieves the state of the art results.

	{\small
		\bibliographystyle{ieee}
		\bibliography{egbib}
	}
	
\end{document}


	\title{LIP: Learning Instance Propagation for Video Object Segmentation\\
	Supplementary Material}
	\author{Ye Lyu\\
		University of Twente\\
		{\small y.lyu@utwente.nl}
		\and
		George Vosselman\\
		University of Twente\\
		{\small george.vosselman@utwente.nl}
		\and
		Gui-Song Xia\\
		Wuhan University\\
		{\small guisong.xia@whu.edu.cn}
		\and
		Michael Ying Yang\\
		University of Twente\\
		{\small michael.yang@utwente.nl}
	}

\maketitle
\ificcvfinal\thispagestyle{empty}\fi
	

\noindent
This supplementary material provides additional details of training, quantitative results, and qualitative examples.

\section{More details of training}

\noindent
\textbf{Multi-level RPN}
5 RPN heads are created for 5 levels of the Conv-GRU module. Each RPN head proposes boxes for 1 scale of all 5 scales.

\noindent
\textbf{Multi-level ROIs}
ROI aligned features are extracted from 1 of the 4 bottom levels of the Conv-GRU module based on box scales ($<112^2$, $112^2-224^2$, $224^2-448^2$, $>448^2$).

\noindent
\textbf{Focal loss}
Focal loss hyper-parameters are set the same as in ~\cite{focalloss}.

\noindent
\textbf{Early stopping}
If the loss for a prediction head is smaller than an empirically chosen threshold, the loss is ignored. If all the losses are ignored, we stop the training.
\textit{Online training:}
The thresholds for bbox regression loss, id cross-entropy loss and mask binary cross-entropy loss are 0.015, 0.010 and 0.15 respectively.
\textit{Online inference:}
The early stopping threshold for id cross-entropy loss is set to 0.015.

\section{Results}

	\subsection{Ablation study}
	We further examine the prediction without \textit{one maximum constraint} and online fine-tuning during inference to see the behavior of the learned model. Example is shown in Fig.~\ref{fig:box_no_suppr}. Most of the detected boxes are around the target objects, except a few boxes covering similar objects nearby. It shows that our model is able to track multiple target objects. It could also be seen that there is decay for probability over time even though the maximum probability is correctly matched to target objects. Such limitation is caused by the fact that the id classifier is learned from the first image, which does not generalize for all images and requires online fine-tuning to preserve probability.
		\begin{figure*}[htbp]
		\begin{center}
			\includegraphics[width=1.0\linewidth]{imgs/box_no_suppr.png}
		\end{center}
		\caption{A qualitative example of prediction without one maximum constraint and online fine-tuning for id head.}
		\label{fig:box_no_suppr}
	\end{figure*}
	
	\subsection{Failed cases and limitation}
	During online inference, we find two failed cases that are most relevant to our method. Although long-term visual memory is included in our model, it still fails to handle some occlusions as shown in Fig.~\ref{fig:fail_case1}. Our model also finds it difficult to infer for multiple instances with large bounding box overlaps as it is difficult to predict the correct ids for all bounding boxes. Example is shown in Fig.~\ref{fig:fail_case2}.
	
	\subsection{Compare with visual memory based method}
	In addition, we compare our method with another visual memory (Conv-GRU) based video object segmentation method~\cite{VOSVisualMemory} on DAVIS 2016 dataset. Both of the methods are trained with additional image dataset, but we achieve better result without optical flow and CRF post-processing, as shown in Table \ref{tab:cmp_mem}. The result shows the value of the learned concept of instances.
	\begin{table}[htp]
		\begin{center}
			\begin{tabular}{r|c|c}
				\hline
				Method              & VisualMem~\cite{VOSVisualMemory} &  LIP(Ours)\\
				\hline
				Additional dataset  &PASCAL VOC&Ms COCO\\
				\hline
				Additional aid      &CRF\&Optical Flow& -\\
				\hline
				J\&F Mean$\uparrow$ & 74.0	 & 78.5\\
				\hline
				J Mean$\uparrow$    & 75.9   & 78.0\\
				\hline
				F Mean$\uparrow$    & 72.1   & 79.0\\
				\hline
			\end{tabular}
		\end{center}
		\caption{Comparison with another visual memory based method~\cite{VOSVisualMemory}. Results reported on DAVIS 2016.}
		\label{tab:cmp_mem}
	\end{table}
	
	\subsection{More qualitative results comparison}
	Here we show more qualitative results comparison of OnAVOS~\cite{OnAVOS}, OSVOS~\cite{OSVOS}, FAVOS~\cite{fast_track_vos}, OSMN~\cite{efficient_vos} and LIP on DAVIS 2016 and DAVIS 2017 datasets in Fig.~\ref{fig:davis16_supp_1},~\ref{fig:davis16_supp_2} and Fig.~\ref{fig:davis17_supp_1},~\ref{fig:davis17_supp_2}.

	\begin{figure*}[htbp]
		\begin{center}
			\includegraphics[width=1.0\linewidth]{imgs/fail1.png}
		\end{center}
		\caption{Example of failed cases due to occlusions.}
		\label{fig:fail_case1}
	\end{figure*}
	
	\begin{figure*}[htbp]
		\begin{center}
			\includegraphics[width=1.0\linewidth]{imgs/fail2.png}
		\end{center}
		\caption{Example of failed cases due to large overlaps between target objects.}
		\label{fig:fail_case2}
	\end{figure*}
	
	\begin{figure*}[htbp]
		\begin{center}
			\includegraphics[width=1.0\linewidth]{imgs/fig_supp_3.png}
		\end{center}
		\caption{Qualitative results comparison of OnAVOS~\cite{OnAVOS}, OSVOS~\cite{OSVOS}, FAVOS~\cite{fast_track_vos}, OSMN~\cite{efficient_vos} and LIP on DAVIS 2016 dataset. The index of each image in a sequence is shown on the top.}
		\label{fig:davis16_supp_1}
	\end{figure*}
	
	\begin{figure*}[htbp]
		\begin{center}
			\includegraphics[width=1.0\linewidth]{imgs/fig_supp_4.png}
		\end{center}
		\caption{Qualitative results comparison of OnAVOS~\cite{OnAVOS}, OSVOS~\cite{OSVOS}, FAVOS~\cite{fast_track_vos}, OSMN~\cite{efficient_vos} and LIP on DAVIS 2016 dataset. The index of each image in a sequence is shown on the top.}
		\label{fig:davis16_supp_2}
	\end{figure*}
	
	\begin{figure*}[htbp]
		\begin{center}
			\includegraphics[width=1.0\linewidth]{imgs/fig_supp_1.png}
		\end{center}
		\caption{Qualitative results comparison of OnAVOS~\cite{OnAVOS}, OSVOS~\cite{OSVOS}, FAVOS~\cite{fast_track_vos}, OSMN~\cite{efficient_vos} and LIP on DAVIS 2017 dataset. The index of each image in a sequence is shown on the top.}
		\label{fig:davis17_supp_1}
	\end{figure*}
	
	\begin{figure*}[htbp]
		\begin{center}
			\includegraphics[width=1.0\linewidth]{imgs/fig_supp_2.png}
		\end{center}
		\caption{Qualitative results comparison of OnAVOS~\cite{OnAVOS}, OSVOS~\cite{OSVOS}, FAVOS~\cite{fast_track_vos}, OSMN~\cite{efficient_vos} and LIP on DAVIS 2017 dataset. The index of each image in a sequence is shown on the top.}
		\label{fig:davis17_supp_2}
	\end{figure*}

	{\small
		\bibliographystyle{ieee}
		\bibliography{egbib}
	}